\documentclass[11pt,a4paper]{article}
\usepackage{hyperref}
\usepackage{acl2018}
\usepackage{times}
\usepackage{latexsym}
\usepackage{amsfonts}
\usepackage{amsmath}
\usepackage{graphicx}
\usepackage{amssymb}

\usepackage{url}

\title{Joint Training of Candidate Extraction and Answer Selection \\for Reading Comprehension}

\author{Zhen Wang\quad Jiachen Liu \quad Xinyan Xiao \quad Yajuan Lyu \quad Tian Wu\\
Baidu Inc., Beijing, China\\
\{wangzhen24, liujiachen, xiaoxinyan, lvyajuan, wutian\}@baidu.com}

\date{}

\begin{document}
\pagestyle{empty}
\maketitle
\begin{abstract}
   While
   sophisticated neural-based techniques have been 
   developed in 
   reading comprehension, most approaches 
   model the answer in an independent manner, 
   ignoring its relations with other answer candidates.
   This problem can be even worse in open-domain scenarios, where 
   candidates from multiple passages should be combined to answer a single question. In this paper, we formulate reading comprehension as an extract-then-select two-stage procedure.
   We first extract answer candidates from passages, then select the final answer 
   by combining information from all the candidates. Furthermore, we
   regard candidate extraction as a latent variable and train the two-stage process jointly with reinforcement learning. 
   As a result, our approach has improved the state-of-the-art performance significantly on two challenging open-domain reading comprehension datasets.
   Further analysis demonstrates the effectiveness of 
   our model components,
   especially the information fusion of all the 
   candidates and the joint training of the extract-then-select procedure.
\end{abstract}

\section{Introduction}
Teaching machines to read and comprehend human languages is a long-standing 
objective in natural language processing.
In order to evaluate this ability, 
reading comprehension (RC) is designed to answer questions 
through reading relevant passages. 
In recent years, RC has attracted intense interest.  Various advanced neural models have been proposed along with newly released datasets \cite{hermann2015teaching, rajpurkar-EtAl:2016:EMNLP2016, dunn2017searchqa, dhingra2017quasar, he2017dureader}.

Most existing approaches mainly focus on modeling the interactions between questions and passages \cite{dhingra-EtAl:2017:Long2, seo2016bidirectional, wang2017gated}, 
paying less attention to information concerning answer candidates.
However, when human 
solve this problem, we often first read each piece of text, collect some answer candidates, then focus on these candidates and combine their
information to select
the final answer. This collect-then-select process can be more significant 
in open-domain scenarios, 
which require the combination of candidates from multiple passages 
to answer one single question.
 This phenomenon is illustrated by the example in Table \ref{intro}.


\begin{table}[t!]
\renewcommand\arraystretch{1.2}
\small
\begin{center}
\begin{tabular}{|p{0.3cm}|p{6.5cm}|}
\hline
{Q} & {Cocktails: \textit{\textcolor{blue}{Rum}}, \textit{\textcolor{blue}{lime}}, and \textit{\textcolor{blue}{cola}} drink make a \_\_\_\_\_\_\_\_\_\_\_\_.} \\
{A} & {{\textcolor{red}{\textbf{Cuba Libre}}}} \\
\hline
$\text P_{\text 1}$ & {\textbf{Daiquiri}, the custom of mixing \textit{\textcolor{blue}{lime}} with \textit{\textcolor{blue}{rum}} for a cooling drink on a hot Cuban day, has been around a long time.} \\
\hline
$\text P_{\text 2}$ & {Cocktail recipe for a \textbf{Daiquiri}, a classic \textit{\textcolor{blue}{rum}} and \textit{\textcolor{blue}{lime}} drink that every bartender should know.}\\
\hline
$\text P_{\text 3}$ & {Hemingway Special \textbf{Daiquiri}: Daiquiris are a family of cocktails whose main ingredients are \textit{\textcolor{blue}{rum}} and \textit{\textcolor{blue}{lime}} juice.}\\
\hline
$\text P_{\text 4}$ & {A homemade Cuba Libre Preparation To make a \textbf{Cuba Libre} properly, fill a highball glass with ice and half fill with \textit{\textcolor{blue}{cola}}.}\\
\hline
$\text P_{\text 5}$ & {The difference between the \textbf{Cuba Libre} and \textit{\textcolor{blue}{Rum}} is a \textit{\textcolor{blue}{lime}} wedge at the end.} \\
\hline
\end{tabular}
\end{center}
\caption{\label{intro}
The answer candidates are in a \textbf{bold} font. The key information is marked in {\it{italic}}, which should be combined from different text pieces to select the correct answer "Cuba Libre".}
\end{table}

With this motivation, we formulate an extract-then-select two-stage architecture to 
simulate the above procedure.
The architecture contains two components: (1) an
extraction model, which generates answer candidates, (2) a selection model, which combines all these candidates and 
finds out the final answer. However, 
answer candidates to be focused on are often unobservable, as most RC datasets only provide golden answers.
Therefore, we treat candidate extraction as a latent variable and train these two stages jointly with reinforcement learning (RL). 

In conclusion, our work makes the following contributions:

1. We formulate open-domain reading comprehension as a two-stage procedure, which first extracts answer candidates and then selects the final answer. With joint training, we 
optimize these two correlated stages as a whole.

2. We propose a novel answer selection model, which combines the 
information from all the extracted candidates using an attention-based correlation matrix.  
As shown in experiments, the information fusion is greatly helpful for answer selection.

3. With the two-stage framework and the joint training strategy, our method significantly surpasses the state-of-the-art performance on two challenging public RC datasets Quasar-T \cite{dhingra2017quasar} and SearchQA \cite{dunn2017searchqa}.


\section{Related Work}
In recent years, reading comprehension has made remarkable progress in methodology and dataset construction.
Most existing approaches mainly focus on modeling sophisticated interactions between questions and passages, then use the pointer networks \cite{vinyals2015pointer} to directly model the answers \cite{dhingra-EtAl:2017:Long2, wang2016machine, seo2016bidirectional, wang2017gated}. These methods prove to be
 effective in existing close-domain datasets \cite{hermann2015teaching, hill2015goldilocks, rajpurkar-EtAl:2016:EMNLP2016}.

More recently, open-domain RC has attracted increasing attention \cite{nguyen2016ms, dunn2017searchqa, dhingra2017quasar, he2017dureader} and raised new challenges for question answering techniques. In these scenarios, a question is paired with multiple passages, which are often collected by exploiting unstructured documents or web data. Aforementioned approaches 
often rely on recurrent neural networks and sophisticated 
attentions, which are prohibitively time-consuming if passages are concatenated altogether. 
 Therefore, some work tried to alleviate this problem in a coarse-to-fine schema. 
\newcite{wang2017r} combined a ranker for selecting the relevant passage and a reader for producing the answer from it. However, this approach only depended on one passage when producing the answer, hence 
put great demands on the precisions of both components.
Worse still, 
this framework cannot handle the situation where multiple passages are needed to answer correctly. 
In consideration of evidence aggregation, \newcite{wang2017evidence} proposed a re-ranking method to 
resolve the above issue. However, their re-ranking stage was totally isolated from the candidate extraction procedure.
Being different from the re-ranking perspective, we propose a novel selection model to combine the information from all the extracted candidates. Moreover, with reinforcement learning, our candidate extraction and answer selection models can be learned in a joint manner. 
\newcite{trischler2016natural} also proposed a two-step extractor-reasoner model, which first extracted $K$ most probable single-token answer candidates and then compared the hypotheses with all the sentences in the passage. However, in their work, each candidate was considered isolatedly, and their objective only took into account the ground truths compared with our RL treatment.

The training strategy employed
in our paper is reinforcement learning, which is inspired by recent work exploiting 
it into question answering problem. The above mentioned coarse-to-fine framework \cite{choi-EtAl:2017:Long, wang2017r} treated sentence selection as a latent variable and jointly trained the sentence selection module with the answer generation module via RL.
\newcite{shen2017reasonet} modeled the multi-hop reasoning procedure with a termination state to decide when it is adequate to produce an answer. RL is suitable to capture this stochastic behavior. \newcite{hu2017reinforced} merely modeled the extraction process, using F1 as rewards 
in addition to maximum likelihood estimation. RL was utilized in their training process, as the F1 measure is not differentiable.

\section{Two-stage RC Framework}
\begin{figure}[t]
  \centering
  \includegraphics[width=0.45\textwidth]{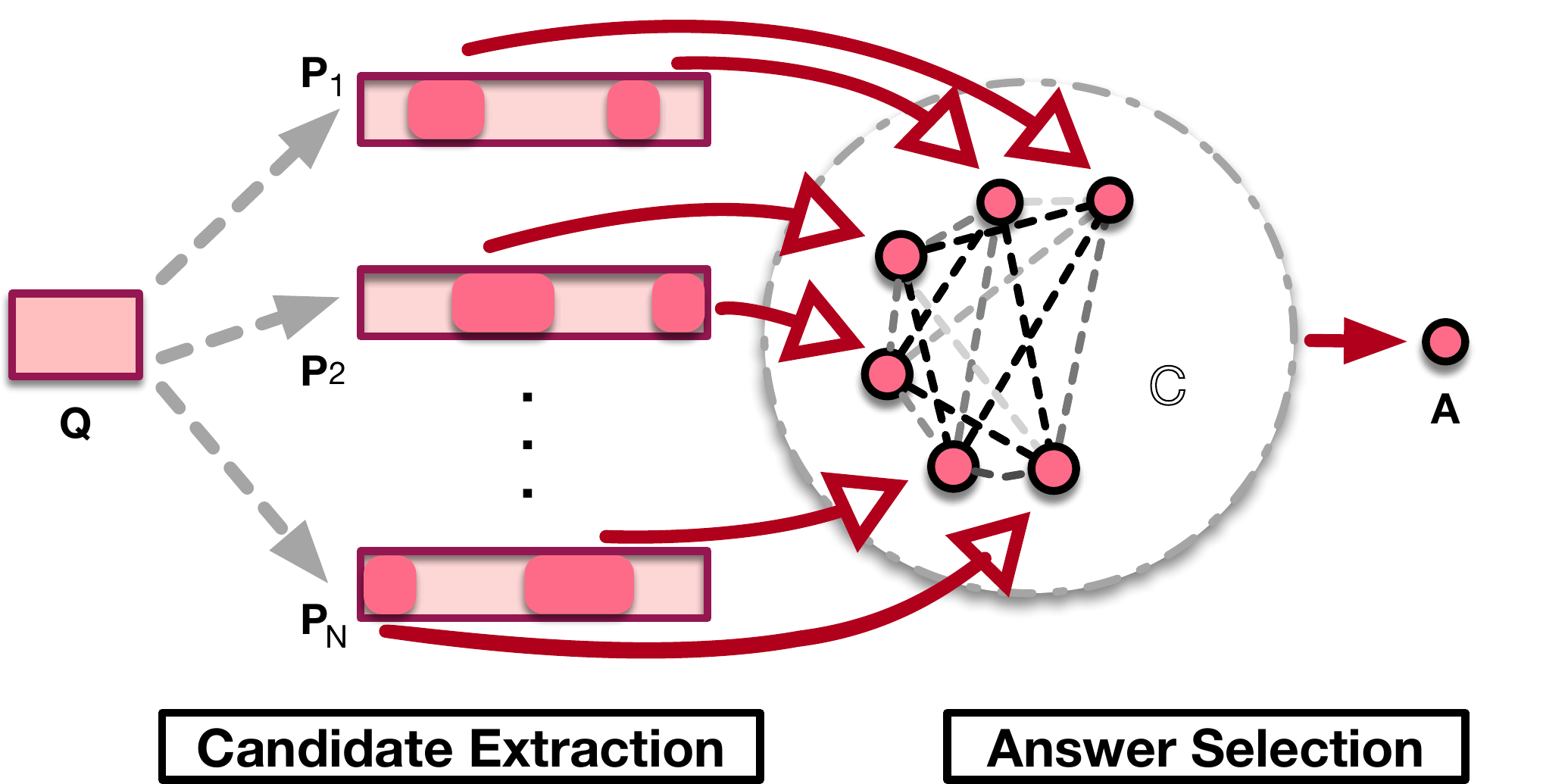}
  \caption{Two-stage RC Framework. The first part extracts candidates (denoted with circles) from all the passages. The second part establishes interactions among all these candidates to select the final answer. The different gray scales of dashed lines between candidates represent different intensities of interactions.}\label{fig:figure1}
\end{figure}
In this work, we mainly consider the open-domain extractive reading comprehension. In this scenario, a given question $Q$ is paired with multiple passages $\mathbb{P}=\{P_{1}, P_{2}, ..., P_{N}\}$,
based on which we aim to find out the answer {\emph A}. Moreover, the golden answers are almost subspans shown in some passages in $\mathbb{P}$. 
Our main framework consists of two parts, which are: (1) extracting answer candidates $\mathbb{C} = \{C_{1}, C_{2}, ..., C_{M}\}$ from passages $\mathbb{P}$
and (2) selecting the final answer $A$ from candidates $\mathbb{C}$. This process is illustrated in Figure \ref{fig:figure1}. We design different models for each part and 
optimize them as a whole with joint reinforcement learning.

\subsection{Candidate Extraction}
We build candidate set $\mathbb C$ by independently extracting $K$ candidates from each passage $P_i$ according to the following distribution:
\begin{equation}
\begin{aligned}
p(\mathbb{C}|Q, \mathbb{P})=&
\prod_{i}^{N}p(\{C_{ij}\}_{j=1}^{{K}}|Q, P_{i})\\
\mathbb{C}=&\bigcup_{i=1}^N \{C_{ij}\}_{j=1}^K
\end{aligned}
\end{equation}
where $C_{ij}$  denotes the $j$th candidate extracted from the $i$th passage. 
${K}$ is set as a constant number in our formulation.
Taking ${K}$ as 2 for an example, we 
denote each probability shown on the right side of Equation 1 through sampling without replacement:
\begin{equation}
\begin{aligned}
p(\{C_{i1}&, C_{i2}\}) =p(C_{i1})p(C_{i2}) / (1 - p(C_{i1})) \\
 &+ p(C_{i1})p(C_{i2}) / (1 - p(C_{i2}))
\end{aligned}
\end{equation}
where we neglect $Q$, $P_{i}$ 
to abbreviate the conditional distributions in Equation 1.

\begin{figure}[t]
  \centering
  \includegraphics[width=0.45\textwidth]{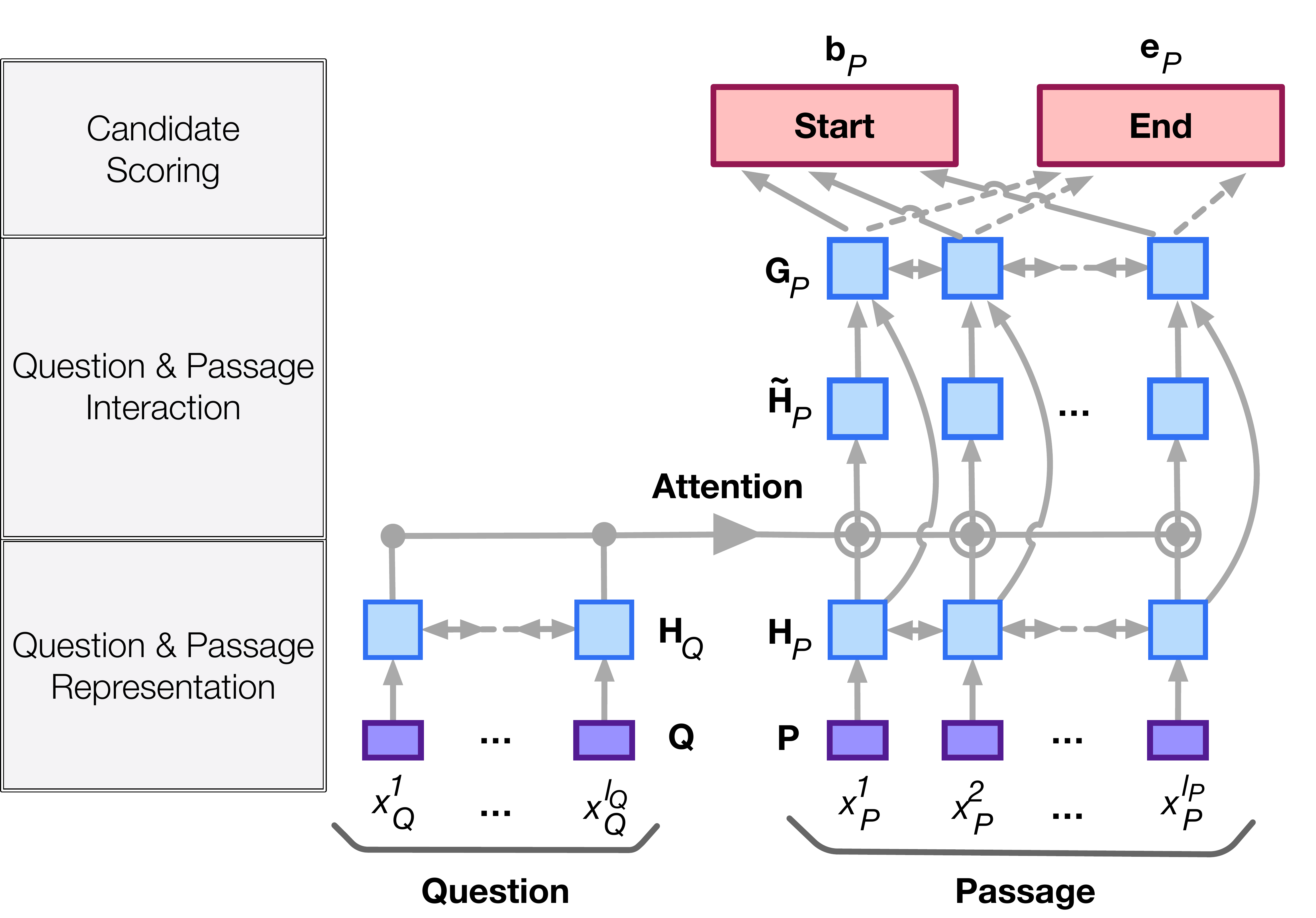}
  \caption{Candidate Extraction Model Architecture.}\label{fig:figure2}
\end{figure}

Consequently, the basic block of our candidate extraction stage 
turns out to be 
 the distribution of each candidate $P(C_{ij}|Q, P_i)$. 
In the rest of this subsection, we will 
elaborate on the model architecture concerning 
candidate extraction, which is displayed in Figure \ref{fig:figure2}.

\paragraph{Question \& Passage Representation}
Firstly, we embed the question $Q=\{x_Q^k\}_{k=1}^{l_Q}$ and its relevant passage $P=\{x_P^t\}_{t=1}^{l_P} \in \mathbb{P}$ with word vectors to form $\textbf{Q}\in \mathbb{R}^{d_{w}\times l_{Q}}$ and $\textbf{P}\in \mathbb{R}^{d_{w}\times l_{P}}$ respectively, where $d_{w}$ is the dimension of word embeddings, $l_{Q}$ and $l_{P}$ are the length of $Q$ and $P$.

We then feed $\textbf{Q}$ and $\textbf{P}$ to a 
bidirectional LSTM to form their contextual representations $\textbf{H}_{Q}\in \mathbb{R}^{d_{h}\times l_{Q}}$ and $\textbf{H}_{P}\in \mathbb{R}^{d_{h}\times l_{P}}$:
\begin{equation}
\begin{aligned}
\textbf{H}_{Q}&={\rm BiLSTM}(\textbf{Q}) \\
\textbf{H}_{P}&={\rm BiLSTM}(\textbf{P})
\end{aligned}
\end{equation}
\paragraph{Question \& Passage Interaction}
Modeling 
the interactions between questions and passages is a critical step in reading comprehension. Here, we adopt the attention mechanism similar to \cite{lee2016learning} to generate question-dependent passage representation $\widetilde{\textbf{H}}_{P}$. Assume 
$\textbf{H}_{Q}=\{\textbf{h}_{Q}^{k}\}_{k=1}^{l_{Q}}$, $\textbf{H}_{P}=\{\textbf{h}_{P}^{t}\}_{t=1}^{l_{P}}$
, we have:
\begin{equation}
\begin{aligned}
\alpha_{tk}=\frac{e^{\textbf{h}_{Q}^{k} \cdot \textbf{h}_{P}^{t}}}{\sum_{k=1}^{l_{Q}}e^{\textbf{h}_{Q}^{k}\cdot \textbf{h}_{P}^{t}}}& \quad 1 \leq k \leq l_{Q},  1 \leq t \leq l_{P} \\
\widetilde{\textbf{h}}_{P}^{t}=\sum_{k=1}^{l_{Q}}\alpha_{tk}&\textbf{h}_{Q}^{k} \quad 1 \leq t \leq l_{P} \\
\widetilde{\textbf{H}}_{P}=&\{\widetilde{\textbf{h}}_{P}^{t}\}_{t=1}^{l_{P}}
\end{aligned}
\end{equation}

After concatenating two kinds of passage representations $\textbf{H}_{P}$ and $\widetilde{\textbf{H}}_{P}$, we use another bidirectional LSTM to get the final representation of every position in passage $P$ as $\textbf{G}_{P}\in \mathbb{R}^{d_{g}\times l_{P}}$:
\begin{equation}
\textbf{G}_{{P}}={\rm BiLSTM}([\textbf{H}_{P};\widetilde{\textbf{H}}_{P}])
\end{equation}

\paragraph{Candidate Scoring}
Then we use two linear transformations $\textbf{w}_b \in \mathbb{R}^{1\times d_{g}}$ and $\textbf{w}_e \in \mathbb{R}^{1\times d_{g}}$ to 
calculate the begin and the end scores for each position:
\begin{equation}
\begin{aligned}
\{b_P^t\}_{t=1}^{l_Q} = \textbf{b}_P = \textbf{w}_b\textbf{G}_{P}\\
\{e_P^t\}_{t=1}^{l_Q} = \textbf{e}_P = \textbf{w}_e\textbf{G}_{P}
\end{aligned}
\end{equation}

At last, we model the 
probability of every subspan in passage $P$ as a candidate 
$C=\{x_{P}^t\}_{t=C_b}^{C_e}$ 
 according to its begin and end position:
\begin{equation}
\begin{aligned}
p(C|Q,P)=\frac{exp({b}_P^{C_{b}} + {e}_P^{C_{e}})}{\sum_{k=1}^{l_{P}}\sum_{t=k}^{l_{P}}exp({{b}_P^{k} + {e}_P^{t}})} \\
\end{aligned}
\end{equation}
In this definition, the probabilities of all the valid answer candidates are already normalized.


\subsection{Answer Selection}
\begin{figure}[t]
\centering
  \includegraphics[width=0.45\textwidth]{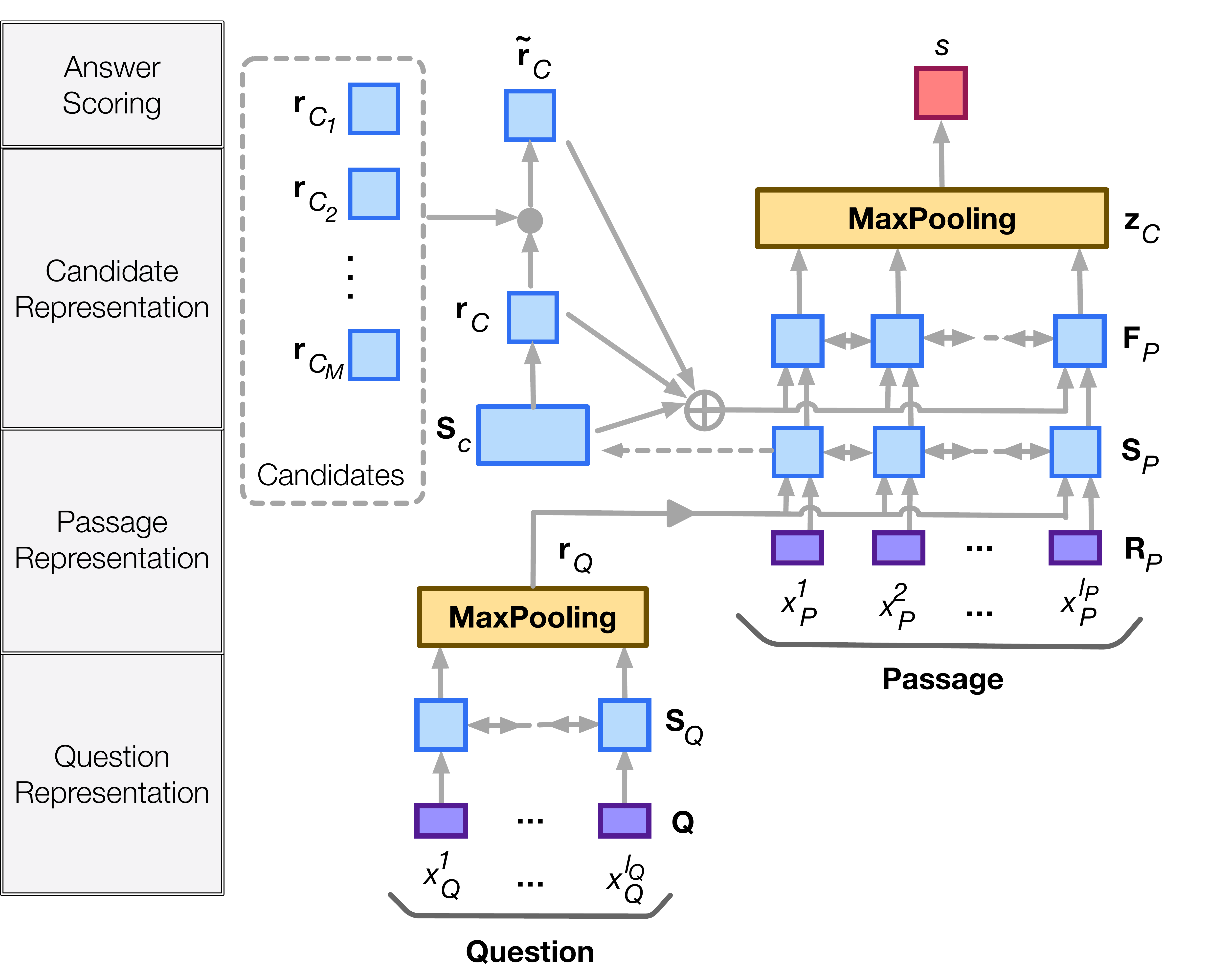}
  \caption{Answer Selection Model Architecture.}\label{fig:figure3}
\end{figure}

As the second part of our framework, the answer selection model finds out the 
most probable answer by calculating $p(C|Q, \mathbb{P}, \mathbb{C})$ for each candidate $C\in\mathbb{C}$. The model architecture is illustrated in Figure \ref{fig:figure3}.

Notably, selection model receives candidate set $\mathbb{C}$ as additional information. This 
more focused information allows the model to 
combine evidences from all the candidates, which would be useful for selecting the best answer. 

For ease of understanding, we briefly describe the selection stage as follows. After being extracted from a single passage, a candidate borrows information from other candidates across different passages. With this global information, the passage is reread to confirm the correctness of the candidate further. The following are details about the selection model.

%

\paragraph{Question Representation}
Questions are 
fundamental for finding out the correct answer.
As did for the extraction model, we embed the question $Q$ with word vectors to form $\textbf{Q}\in \mathbb{R}^{d_{w}\times l_{Q}}$. Then we use a bidirectional LSTM to establish its contextual representation:
\begin{equation}
\begin{aligned}
\textbf{S}_{q}&={\rm BiLSTM}(\textbf{Q}) \\
\end{aligned}
\end{equation}
A max-pooling operation across all the positions is followed to get the condensed vector representation:
\begin{equation}
\begin{aligned}
\textbf{r}_{q} &= {\rm MaxPooling}(\textbf{S}_{q})
\end{aligned}
\end{equation}
\paragraph{Passage Representation}
Assume the candidate $C$ is extracted from the passage $P\in\mathbb{P}$. To be informed of $C$, we first build the representation of $P$.
For every word in $P$,
three kinds of features are utilized:
\begin{itemize}
\item Word embedding: each word expresses its basic feature with the word vector. 
\item Common word: the feature has value 1 when the word occurs in the question, otherwise 0.
\item Question independent representation: the condensed representation $\textbf{r}_{q}$.
\end{itemize}
With these features, 
information not only in 
$Q$ but also in 
$P$ is considered. By concatenating them, we get $\textbf{r}_{P}^{t}$ corresponding to every position $t$ in passage $P$. Then with another bidirectional LSTM, we fuse these features to form the contextual representation of $P$ as $\textbf{S}_{P}\in \mathbb{R}^{d_{s}\times l_{P}}$:
\begin{equation}
\begin{aligned}
&\textbf{R}_{P}=\{\textbf{r}_{P}^{t}\}_{t=1}^{l_{P}} \\
\textbf{S}_{P}&={\rm BiLSTM}(\textbf{R}_{P})
\end{aligned}
\end{equation}

\paragraph{Candidate Representation}
Candidates provide more 
focused information for answer selection. Therefore, for each candidate, we first build its independent representation according to its position in the passage, then construct candidates fused representation through combination of other correlated candidates.

Given the candidate $C=\{x_{P}^t\}_{t=C_b}^{C_e}$ in the passage $P$, we extract its corresponding span from $\textbf{S}_{P}=\{\textbf{s}_{P}^{t}\}_{t=1}^{l_P}$ to form $\textbf{S}_C=\{\textbf{s}_{P}^{t}\}_{t=C_b}^{C_e}$ as its contextual 
encoding.
Moreover, we calculate its condensed vector representation 
through its begin and end positions:
\begin{equation}
\textbf{r}_{C}={\rm tanh}({\textbf W}_{b}\textbf{s}_{P}^{C_{b}}+{\textbf W}_{e}\textbf{s}_{P}^{C_{e}})
\end{equation}
where $\textbf{W}_b \in \mathbb{R}^{d_{c}\times d_{s}}$, $\textbf{W}_e \in \mathbb{R}^{d_{c}\times d_{s}}$.

To model the interactions among all the answer candidates, we calculate the correlations of the candidate $C$, which is assumed to be indexed by $j$ in $\mathbb{C}$, with others 
$\{C_m\}_{m=1, m\neq j}^{M}$ via attention mechanism:
\begin{equation}
V_{jm} = {\textbf w_{v}}{\rm tanh}({\textbf W_{c}}\textbf{r}_{C}+{\textbf W_{o}}\textbf{r}_{C_{m}})
\end{equation}
where $\textbf{W}_c \in \mathbb{R}^{d_{c}\times d_{c}}$, $\textbf{W}_o \in \mathbb{R}^{d_{c}\times d_{c}}$ and $\textbf{w}_v \in \mathbb{R}^{1\times d_{c}}$ are linear transformations to capture the intensity of each interaction.

In this way, we form a {\it correlation matrix} $\textbf V\in \mathbb{R}^{M\times M}$, where $M$ is the total number of candidates.
With the correlation matrix, for the candidate $C$, we normalize its interactions via a $softmax$ operation, which emphasizes the influence of stronger interactions:
\begin{equation}
\alpha_{m}=\frac{e^{V_{jm}}}{\sum_{m=1, m\neq j}^{M}e^{V_{jm}}}
\end{equation}

To take into account different influences of all the other candidates, it is sensible to generate a candidates fused representation 
 according to the above normalized interactions: 
\begin{equation}
\widetilde{\textbf r}_{C}={\sum_{m=1, m\neq j}^{M}\alpha_{m}{\textbf r}_{C_m}}
\end{equation}

In this formulation, all the other candidates contribute their influences to the fused representation by their interactions with $C$, thus information from different passages is gathered altogether. In our experiments, this kind of information fusion is the key point for performance improvements.

\paragraph{Passage Advanced Representation}
As more focused information of the candidate $C$ is available, we are provided with a better way to confirm its correctness by rereading its corresponding passage $P$. Specifically, we equip each position $t$ in $P$ with following advanced features:
\begin{itemize}
\item Passage contextual representation: the former passage representation $\textbf{s}_{P}^{t}$.
\item 
Candidate-dependent passage representation: replace $\textbf{H}_{Q}$ with $\textbf{S}_{C}$ and $\textbf{H}_{P}$ with $\textbf{S}_{P}$ in Equation 4 to model the interactions between candidates and passages to form $\widetilde{\textbf{s}}_{P}^{t}$.

\item Candidate related distance feature: the relative distance to the candidate $C$ can be a reference of the importance of each position.
\item Candidate independent representation: use ${\textbf r}_{C}$ to consider the concerned candidate $C$.
\item Candidates fused representation: use $\widetilde{\textbf r}_{C}$ to consider all the other candidates interacting with the concerned candidate $C$.
\end{itemize}

With these features, we capture the information from the question, the passages and all the candidates. By concatenating them, we get $\textbf{u}_P^t$ in every position in the passage $P$. Combining these features with a bidirectional LSTM, we get:
\begin{equation}
\begin{aligned}
&\textbf{U}_{P}=\{\textbf{u}_{P}^{t}\}_{t=1}^{l_{P}} \\
\textbf{F}_{P}&={\rm BiLSTM}(\textbf{U}_{P})
\end{aligned}
\end{equation}

 
\paragraph{Answer Scoring}
At last, the max pooling of each dimension of $\textbf{F}_{P}$ is performed, resulting in a condensed vector representation, which contains all the concerned information in a candidate:
\begin{equation}
\textbf{z}_C = {\rm MaxPooling}(\textbf{F}_{P})
\end{equation}

The final score of this candidate as the answer is calculated via a linear transformation, which is then normalized across all the candidates:
\begin{equation}
\begin{aligned}
s = \textbf{w}&_{z}\textbf{z}_{C} \\
p(C|Q,\mathbb{P},\mathbb{C}) &= \frac{e^{s}}{\sum_{k=1}^{M}e^{s_{k}}}
\end{aligned}
\end{equation}

\subsection{Joint Training with RL}
In our formulation, the answer candidate set influences the result of answer selection to a large extent.
However, with only golden answers provided in the training data, it is not 
apparent which candidates should be considered further. 

To alleviate the above problem, we treat candidate extraction as a latent variable, jointly train the extraction model and the selection model with reinforcement learning. 
Formally, in the extraction and selection stages, two kinds of actions are modeled. The action space for the extraction model is to select from different candidate sets, which is formulated by Equation 1. The action space for the selection model is to select from all extracted candidates, which is formulated by Equation 17.
 Our goal is to select the final answer that leads to a high reward. Inspired by \newcite{wang2017r}, we define the reward of
a candidate 
to reflect its 
accordance with the golden answer:
\begin{equation}
r(C, A)=\left\{
\begin{array}{ccc}
2         &  & {\rm if} \  C == A\\
f1(C, A)  & &    {\rm else \ if} \  C \cap A \neq \varnothing \\
-1         &  & {\rm{else}}
\end{array} \right.
\end{equation}
where $f1(.,.)\in[0,1]$ is the function to measure word-level F1 score between two sequences. Incorporating this reward can alleviate the overstrict requirements set by traditional maximum likelihood estimation as well as keep consistent with our evaluation methods in experiments.

The learning objective becomes to maximize the expected reward modeled by our framework, where $\theta$ stands for all the parameters involved:
\begin{equation}
\begin{aligned}
L(\theta)&=-E_{\mathbb{C}\sim P(\mathbb{C}|Q,\mathbb{P})}[E_{C\sim P(C|Q,\mathbb{P},\mathbb{C})}r(C, A)]\\
&=-E_{\mathbb{C}\sim P(\mathbb{C}|Q,\mathbb{P})}[\sum_{C}P(C|Q,\mathbb{P},\mathbb{C})r(C, A)]
\end{aligned}
\end{equation}

Following REINFORCE algorithm, we approximate the gradient of the above objective with a sampled candidate set, $\mathbb{C}\sim P(\mathbb{C}|Q,\mathbb{P})$, resulting in the following form:
\begin{equation}
\begin{aligned}
\nabla L(\theta) \approx -\sum_{C}\nabla P(C|Q,\mathbb{P},\mathbb{C})r(C, A) \\
- \nabla logP(\mathbb{C}|Q,\mathbb{P})[\sum_{C}P(C|Q,\mathbb{P},\mathbb{C})r(C, A)]
\end{aligned}	
\end{equation}

\section{Experiments}

\subsection{Datasets}
We evaluate our models on two publicly available open-domain RC datasets, which are commonly adopted in related work.

\textbf{Quasar-T} \cite{dhingra2017quasar} consists of 43,000 open-domain trivia questions and 
corresponding 
answers obtained from various internet sources. Each question is paired with 100 sentence-level passages retrieved from ClueWeb09 
\cite{callan2009clueweb09} based on Lucene. 

\textbf{SearchQA} \cite{dunn2017searchqa} starts from existing question-answer pairs, which are crawled from J!Archive, and is augmented with text snippets retrieved by Google, resulting in more than 140,000 question-answer pairs with each pair having 49.6 snippets on average.
\begin{table}[t!]
\renewcommand\arraystretch{1.2}
\small
\begin{center}
\begin{tabular}{|c|c|c|c|c|c|}
\hline & \#q(train) & \#q(dev) & \#q(test) & \#p\\ \hline
Quasar-T & 28,496 & 3,000 & 3,000 & 100\\
\hline
SearchQA & 99,811 & 13,893 & 27,247 & 50 \\
\hline
\end{tabular}
\end{center}
\caption{\label{datasets} The statistics of our experimental datasets. \#q represents the number of questions for each split of the datasets. \#p is the number of passages for each question.} 
\end{table}

The detailed statistics of these two datasets is shown in Table \ref{datasets}.
\subsection{Model Settings}
We initialize word embeddings with the 300-dimensional Glove vectors\footnote{http://nlp.stanford.edu/data/wordvecs/glove.840B.300d.zip}.
All the bidirectional LSTMs 
hold 1 layer and 100 hidden units. All the linear transformations 
take 
the size of 100 as output dimension.
The common word feature and the candidate related distance feature are embedded with vectors of dimension 4 and 50 respectively. By default, we set $K$ as 2 in Equation 1, which means each passage generates two candidates 
based on the extraction model.

\begin{table*}[t!]
\renewcommand\arraystretch{1.2}
\small
\begin{center}
\begin{tabular}{|l|cccc|}
\hline
\ & \multicolumn{2}{c}{\textbf{Quasar-T}} & \multicolumn{2}{c|}{\textbf{SearchQA}}\\
\ & \textbf{EM} & \textbf{F1} & \textbf{EM} & \textbf{F1}\\
\hline
GA \cite{dhingra-EtAl:2017:Long2}& 26.4 & 26.4 & -& - \\
BIDAF \cite{seo2016bidirectional} & 25.9 & 28.5 & 28.6 & 34.6 \\
AQA \cite{buck2017ask} & - &-&38.7& 45.6 \\
$\rm{R^3}$ \cite{wang2017r} & 35.3 & 41.7 &49.0&55.3\\
\hline
\multicolumn{5}{|l|}{Re-Ranker \cite{wang2017evidence}} \\
\hline
Strength-Based Re-Ranker (Probability) & 36.1 & 42.4&50.4&56.5 \\
Strength-Based Re-Ranker (Counting) & 37.1 & 46.7&54.2&61.6 \\
Coverage-Based Re-Raner & 40.6 & 49.1&53.6&60.6 \\
Full Re-Ranker & 42.3 & 49.6&57.0&63.2 \\
\hline
\multicolumn{5}{|l|}{Our Methods} \\
\hline
Extraction Model & 35.4& 41.6&44.7&51.2\\
Extraction + Selection (Isolated Training) & 41.6& 49.5 &49.7&56.6\\
Extraction + Selection (Joint Training)  &   \textbf{45.9} & \textbf{53.9}& \textbf{58.3} & \textbf{64.2}\\
\hline
\end{tabular}
\end{center}
\caption{\label{quasart-result} Experimental results on the test set of Quasar-T and SearchQA. Full re-ranker is the ensemble of three different re-rankers in \cite{wang2017evidence}.}
\end{table*}

For ease of training, we first initialize our models by maximum likelihood estimation
and fine-tune them with RL.
The similar training strategy is commonly employed when RL process is involved 
 \cite{ranzato2015sequence,li2016deep,hu2017reinforced}. 
To pre-train the extraction model, we only use passages containing ground truths as training data.
The log likelihood of Equation 7 is taken as the training
 objective for each question and passage pair. After pre-training the extraction model, we use it to generate two top-scoring candidates from each passage, forming the training data to pre-train our selection model, and maximize the log likelihood of the Equation 17 as our second objective. In pre-training, we use the batch size of 30 for the extraction model, 20 for the selection model and RMSProp \cite{tieleman2012lecture} with an initial learning rate of 2e-3.
 In fine-tuning with RL, we use the batch size of 5 and RMSProp with an initial learning rate of 1e-4. Also, we use a dropout rate 
 of 0.1 in each training procedure.

\subsection{Experimental Results}
In addition to results of previous work,
we add two 
baselines to 
demonstrate the effectiveness of our framework. The first baseline only applies the extraction model to score the answers, which is aimed at explaining the importance of the
selection model. The second one only uses the pre-trained extraction model and selection model 
to illustrate the benefits from our joint training schema.


The often used evaluation metrics for extractive RC
 are exact match (EM) and F1
\cite{rajpurkar-EtAl:2016:EMNLP2016}. 
The experimental results on Quasar-T and SearchQA are shown in Table \ref{quasart-result}.

As seen from the results on Quasar-T, 
our quite simple extraction model alone almost reaches the state-of-the-art result compared with other methods without re-rankers.
The combination of the extraction and selection models exceeds our extraction baseline by a great margin, and also results in performance surpassing the best single re-ranker in \cite{wang2017evidence}. This result 
illustrates the necessity of introducing the selection model, which incorporates information from all the candidates. In the end, by joint training with RL, our method 
produces better performance even compared with the ensemble of three different re-rankers.

On SearchQA, we find that
 our extraction model alone performs not that well compared with the state-of-the-art model without re-rankers. However, the improvement brought by our selection model isolatedly or jointly trained still demonstrates the importance of our two-stage framework.
 Not surprisingly, comparing the results, our isolated training strategy still lags behind the single re-ranker proposed in 
 \cite{wang2017evidence}, partly because of the deficiency with our extraction model. However, uniting our extraction and selection models with RL makes up the disparity, and the performance surpasses the ensemble of three different re-rankers, let alone the result of any single re-ranker.

\subsection{Further Analysis}



\begin{table}[t!]
\renewcommand\arraystretch{1.2}
\small
\begin{center}
\begin{tabular}{|l|c|c|}
\hline
{\textbf{Quasar-T}} & \ \textbf{EM} & \textbf{F1} \\
\hline
Extraction + Selection (Joint Training)  &   \textbf{45.9} & \textbf{53.9} \\
\hline
-question representation  &   {42.5} & {50.5} \\
-question and passage common words  &   {41.0} & {48.7} \\
-candidate independent representation  &   {44.5} & {53.3} \\
-candidate related distance feature & {44.7} & {53.0} \\
-candidate dependent passage representation  &   {44.4} & {52.3} \\
-candidates fused representation &   {39.2} & {45.8} \\
\hline
\end{tabular}
\end{center}
\caption{\label{ablation} Ablation results concerning the selection model on the test set of Quasar-T. Obviously,  candidates fused representation is the most evident feature when modeling the answer selection procedure.}
\end{table}

\paragraph{Effect of Features in Selection Model}
As the incorporation of the selection model improves the overall performance significantly, we conduct ablation analysis on the Quasar-T 
to 
prove the effectiveness of its major components. 
 As shown in Table \ref{ablation}, all these components modeling the selection procedure play important roles in our final architecture.
 
 Specifically, introducing the independent representation of the question and its common words with the passage seems an efficient way to consider the information of questions,
   which is consistent with previous work
   \cite{li2016dataset,chen-EtAl:2017:Long4}. 
   
   As for features related to candidates, 
   the incorporation of the candidate independent information contributes to the final result more or less. These features include candidate-dependent passage representation, candidate independent representation and candidate related distance feature.  

\begin{table}[t!]
\renewcommand\arraystretch{1.2}
\scriptsize
\begin{center}
\begin{tabular}{|p{0.3cm}|p{6.5cm}|}
\hline
Q & {Cocktails : Rum , lime , and cola drink make a \_\_\_\_\_\_\_\_\_\_\_\_ .} \\
A & {{\textcolor{red}{\textbf{Cuba Libre}}}} \\
\hline
$\rm P_{1}$ & {In Nicaragua , when it is mixed using Flor de Ca a -LRB- the national brand of rum -RRB- and cola , it is called a \textbf{Nica Libre} .} \\
$\rm P_{2}$ & {The drink ... \textbf{Daiquiri} The custom of mixing lime with rum for a cooling drink on a hot Cuban day has been around a long time .} \\
$\rm P_{3}$ & {If you only learn to make two cocktails , the \textbf{Manhattan} should be one of them .}\\
$\rm P_{4}$ & {\textbf{Daiquiri} Cocktail recipe for a Daiquiri , a classic rum and lime drink that every bartender should know .}\\
$\rm P_{5}$ & {Hemingway Special \textbf{Daiquiri} : Daiquiris are a family of cocktails whose main ingredients are rum and lime juice .}\\
$\rm P_{6}$ & {In the Netherlands the drink is commonly called \textbf{Baco} , from the two ingredients of Bacardi rum and cola .}\\
$\rm P_{7}$ & {A homemade Cuba Libre Preparation To make a \textbf{Cuba Libre} properly , fill a highball glass with ice and half fill with cola .}\\
$\rm P_{8}$ & {\textbf{Bacardi} Cocktail Cocktail recipe for a Bacardi Cocktail , a classic cocktail of Bacardi rum , lemon or lime juice and grenadine Roy Rogers -LRB- non-alcoholic -RRB- Cocktail recipe for a Roy Rogers ,}\\
$\rm P_{9}$ & {\textbf{Margarita} Cocktail recipe for a Margarita , a popular refreshing tequila and lime drink for summer .} \\
$\rm P_{10}$ & {The difference between the \textbf{Cuba Libre} and Rum is a lime wedge at the end .} \\
\hline
\end{tabular}
\end{center}
\caption{\label{example} An example from Quasar-T to illustrate the necessity of fused information. Candidates extracted from passages are in a \textbf{bold} font. To correctly answer the question, information in $\rm P_{7}$ and $\rm P_{10}$ should be combined. }
\end{table}

\begin{figure}[!t]
  \centering
  \includegraphics[width=0.5\textwidth]{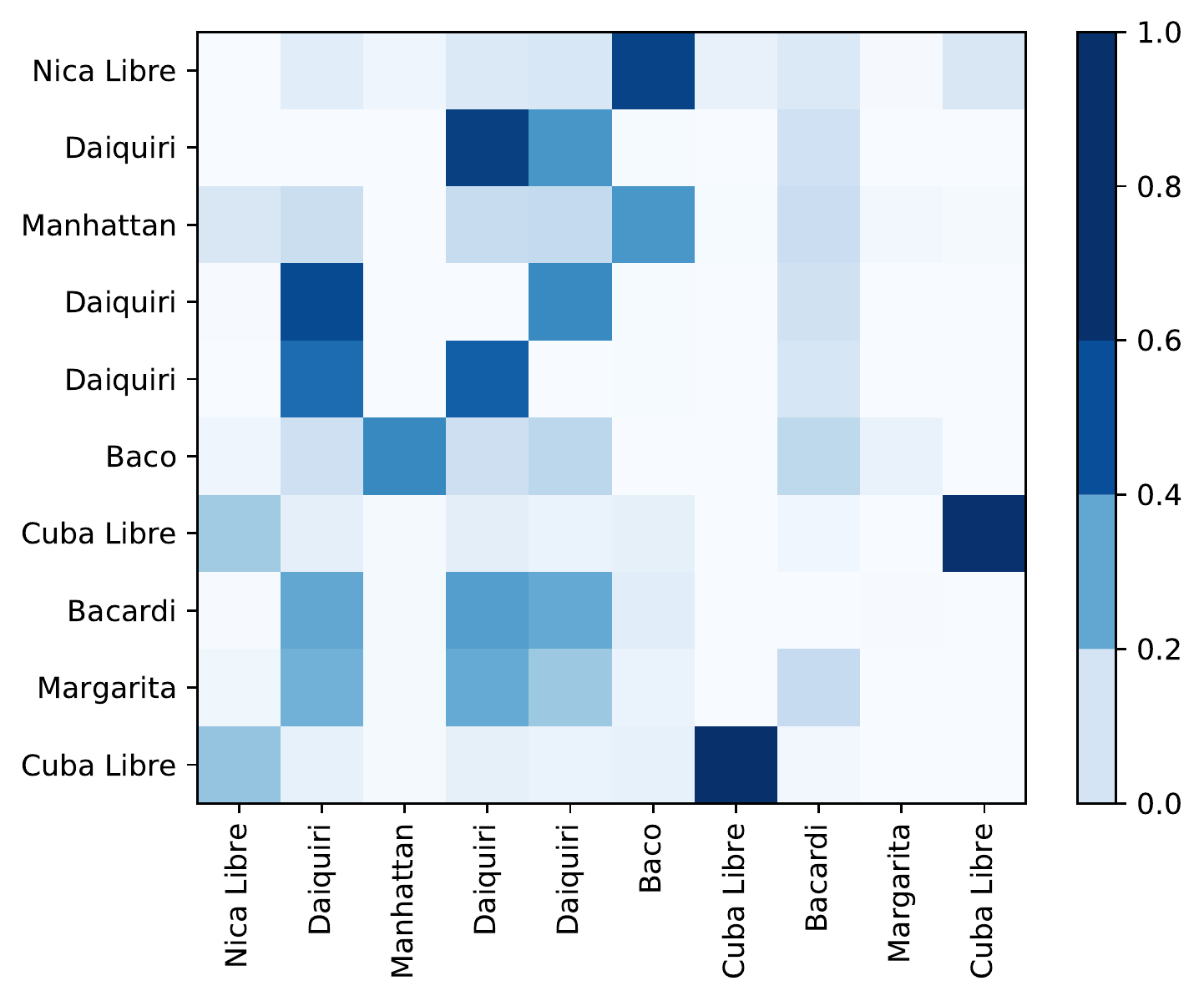}
  \caption{The attention map generated when modeling candidates fused representations for the example in Table \ref{example}.}\label{fig:figure5}
\end{figure}

Most importantly, the candidates fused representation, which combines the information from all the candidates, demonstrates its indispensable role in candidate modeling, with a performance drop of nearly 8\% when discarded. This phenomenon also verifies the necessity of our extract-then-select procedure, showing the importance of combining information scattered in different text pieces when picking out the final answer. 
\paragraph{Example for Candidates Fused Representation}
   We conduct a case study to 
   demonstrate the importance of candidates fused information further. In Table \ref{example}, each candidate only partly matches the description of the question in its independent context. To correctly answer the question, information in $\rm P_{7}$ and $\rm P_{10}$ should be combined. In experiments, our selection model provides the correct answer, while the wrong candidate "Daiquiri", a different kind of cocktail, is selected if candidates fused representation is discarded. The attention map established when modeling the fusion of candidates (corresponding to Equation 13) in this example is illustrated in Figure \ref{fig:figure5}, in which we can see the interactions among all the candidates from different passages. In this figure, it is obvious that 
   the interaction of "Cuba Libre" in $\rm P_7$ and $\rm P_{10}$ is the key point to answer the question correctly.

\paragraph{Effect of Candidate Number}
The candidate extraction stage 
takes an important role to decide what information should be focused on further.
Therefore, we also test the influence of different $K$ when extracting candidates from each passage. The results 
are shown in Table \ref{k-analysis}. Taking $K=1$ degrades the performance, which conforms to the expectation, as the correct candidates become less in this stricter situation. However, taking $K=3$ can not improve the performance further. Although a larger $K$ means 
a higher possibility to include good answers, it 
raises more challenges for the selection model to pick out the correct one from candidates with more varieties. 

\begin{table}[!t]
\renewcommand\arraystretch{1.2}
\small
\begin{center}
\begin{tabular}{|c|c|c|}
\hline
{\textbf{Quasar-T}} & \ \textbf{EM} & \textbf{F1} \\
\hline
\hline
K=1  &   {43.9} & {52.4} \\
K=2 &   {45.9} & {53.9} \\
K=3  &   {45.8} & {53.9} \\
\hline
\end{tabular}
\end{center}
\caption{\label{k-analysis} Different number of extracted candidates results in different final performance on the test set of Quasar-T.}
\end{table}

\section{Conclusion}
In this paper, we formulate the problem of 
RC as a two-stage process, which first generates candidates with an extraction model, then selects the final answer by combining the information from all the candidates. 
Furthermore, we treat candidate extraction as a latent variable and jointly train these two stages with RL.
    Experiments on public open-domain RC datasets Quasar-T and SearchQA show the necessity of introducing the selection model and the effectiveness of fusing candidates information when modeling. Moreover, our joint training strategy leads to significant improvements in performance. 

\section*{Acknowledgments}
This work is supported by the National Basic Research Program of China (973 program, No. 2014CB340505). We thank Ying Chen and
anonymous reviewers for valuable feedback.
\bibliography{acl2018}
\bibliographystyle{acl_natbib}
\end{document}